\documentclass{article}

\usepackage{microtype}
\usepackage{graphicx}
\usepackage{subfigure}
\usepackage{booktabs} % for professional tables

\usepackage{multirow}
\usepackage{bbding}
\usepackage{amsmath}

\usepackage{hyperref}

\usepackage[accepted]{icml2025}

\usepackage{amsmath}
\usepackage{amssymb}
\usepackage{mathtools}
\usepackage{amsthm}
\usepackage{upgreek}

\usepackage[capitalize,noabbrev]{cleveref}

\theoremstyle{plain}

\theoremstyle{definition}

\theoremstyle{remark}

\usepackage[textsize=tiny]{todonotes}

\icmltitlerunning{No Training Wheels: Steering Vectors for Bias Correction at Inference Time}

\begin{document}

\twocolumn[
\icmltitle{No Training Wheels: \\ Steering Vectors for Bias Correction at Inference Time}

% It is OKAY to include author information, even for blind
% submissions: the style file will automatically remove it for you
% unless you've provided the [accepted] option to the icml2025
% package.

% List of affiliations: The first argument should be a (short)
% identifier you will use later to specify author affiliations
% Academic affiliations should list Department, University, City, Region, Country
% Industry affiliations should list Company, City, Region, Country

% You can specify symbols, otherwise they are numbered in order.
% Ideally, you should not use this facility. Affiliations will be numbered
% in order of appearance and this is the preferred way.
\icmlsetsymbol{equal}{*}

\begin{icmlauthorlist}
\icmlauthor{Aviral Gupta}{equal,yyy}
\icmlauthor{Armaan Sethi}{equal,yyy}
\icmlauthor{Ameesh Sethi}{yyy}
\end{icmlauthorlist}

\icmlaffiliation{yyy}{BITS Pilani, Pilani, India}
% \icmlaffiliation{comp}{Company Name, Location, Country}
% \icmlaffiliation{sch}{School of ZZZ, Institute of WWW, Location, Country}

\icmlcorrespondingauthor{Aviral Gupta}{f20220097@pilani.bits-pilani.ac.in}
% \icmlcorrespondingauthor{Firstname2 Lastname2}{first2.last2@www.uk}

% You may provide any keywords that you
% find helpful for describing your paper; these are used to populate
% the "keywords" metadata in the PDF but will not be shown in the document
\icmlkeywords{Bias Mitigation, Fairness, AI Safety, Machine Learning, ICML}

\vskip 0.3in
]

% this must go after the closing bracket ] following \twocolumn[ ...

% This command actually creates the footnote in the first column
% listing the affiliations and the copyright notice.
% The command takes one argument, which is text to display at the start of the footnote.
% The \icmlEqualContribution command is standard text for equal contribution.
% Remove it (just {}) if you do not need this facility.

%\printAffiliationsAndNotice{}  % leave blank if no need to mention equal contribution
\printAffiliationsAndNotice{\icmlEqualContribution} % otherwise use the standard text.

\begin{abstract}
Neural network classifiers trained on datasets with uneven group representation often inherit class biases and learn spurious correlations. These models may perform well on average but consistently fail on atypical groups. For example, in hair color classification, datasets may over-represent females with blond hair, reinforcing stereotypes. Although various algorithmic and data-centric methods have been proposed to address such biases, they often require retraining or significant compute. In this work, we propose a cheap, training-free method inspired by steering vectors used to edit behaviors in large language models. We compute the difference in mean activations between majority and minority groups to define a "bias vector," which we subtract from the model’s residual stream. This leads to reduced classification bias and improved worst-group accuracy. We explore multiple strategies for extracting and applying these vectors in transformer-like classifiers, showing that steering vectors, traditionally used in generative models, can also be effective in classification. More broadly, we showcase an extremely cheap, inference time, training free method to mitigate bias in classification models.
\end{abstract}

\section{Introduction}
\label{intro}

Modern classification models have achieved remarkable success in tasks across a variety of domains, ranging from facial attribute recognition to medical diagnostics and natural language understanding. Despite their success, these models are known to suffer from representational biases when trained on datasets with skewed demographic distributions. These models often inherit and amplify societal biases, because of their reliance on spurious correlations learned during training, which often leads to degraded performance on underrepresented or atypical groups \citep{geirhos2020shortcut, sagawa2019distributionally}. For instance, in facial recognition datasets, attributes like hair color may be disproportionately represented across genders, causing models to associate certain features with specific groups, thereby reinforcing stereotypes. Such biases not only undermine the fairness of AI systems but also pose significant challenges in deploying models in real-world, diverse environments.

Traditional approaches to mitigate these biases include re-sampling techniques, adversarial training, and the incorporation of fairness constraints during model optimization. While effective to some extent, these methods often require extensive retraining, access to sensitive group labels, or complex modifications to the training pipeline. Consequently, there's a growing need for lightweight, post-hoc techniques that can adjust model behaviors without the overhead of retraining or the necessity of sensitive data.

In the realm of large language models (LLMs), the technique of activation steering has emerged as a powerful method for modulating model behaviors at inference time without necessitating retraining. Based on the hypothesis that LLMs represent features or concepts as linear directions in activation space \citep{bolukbasi2016man, park2023linear}, this approach involves identifying and manipulating specific directions, termed steering vectors, in the model's activation space to adjust specific behaviors and attributes in the model's generations. By computing these vectors through contrastive activation differences between inputs exhibiting opposing characteristics \citep{panickssery2024steeringllama2contrastive, burns2024discoveringlatentknowledgelanguage}, researchers have successfully been able to extract feature directions for precise concepts and control model behaviors related to the same.

Building upon these insights, our work explores the novel application of steering vectors within classification models, a domain where activation steering remains underexplored. We propose a training-free, inference-time method to mitigate biases arising from uneven group representations in training data. Our approach involves computing mean activation differences between majority and minority groups within a class to derive a "bias vector." By subtracting this vector from the model's residual stream activations during inference, we effectively reduce classification bias. This adjustment leads to improved metrics, such as worst-group accuracy for minority groups, while maintaining overall performance for other groups. Through extensive experimentation across various classification scenarios and steering vector extraction and application choices, we demonstrate the efficacy of this technique, highlighting its potential as a practical tool for enhancing fairness in transformer-based classification models.

Our key contributions are as follows:

\begin{itemize}
\item \textbf{Adaptation of Steering Vectors for Classification Models:} We introduce a training-free, inference-time method to derive and apply steering vectors in classification tasks, targeting the mitigation of biases arising from uneven group representations.
\item \textbf{Comprehensive Evaluation Across Diverse Datasets and Implementation Choices:} We validate our approach on multiple benchmark datasets known for inherent biases, demonstrating significant improvements in fairness metrics without compromising overall accuracy. We also extensively explore various methods of extracting and applying steering vectors, including per layer, all at once, different for each layer etc.
\end{itemize}

Our findings suggest that steering vectors offer a promising avenue for bias mitigation in classification models, providing a scalable and efficient alternative to traditional retraining-based methods.

% \section{Related Work}
% \label{related_work}

\section{Methodology}

\subsection {Preliminaries}
\label{preliminaries}

%%TODO: Add some prelimiary about transformers for classification, ViT and BERT, explaining what the arch is, see transformer section from refusal paper.
\textbf{Transformers for Classification.} Transformer models \citep{vaswani2017attention} employ multi-head self-attention mechanisms and position-wise feedforward layers to model contextual relationships within input sequences. For classification tasks, an input sequence $t = (t_1, t_2, \ldots, t_n) \in V^n$ is first embedded into a sequence of continuous representations $x^{(1)}_i = \text{Embed}(t_i) \in \mathbb{R}^{d_{\text{model}}}$, where $i \in \{1, \ldots, n\}$. These embeddings are then passed through $L$ stacked transformer layers, each consisting of an attention sublayer and a feedforward (MLP) sublayer applied with residual connections:
\begin{align}
    \tilde{x}^{(l)}_i = x^{(l)}_i + \text{Attn}^{(l)}(x^{(l)}_{1:n}), \quad x^{(l+1)}_i = \tilde{x}^{(l)}_i + \text{MLP}^{(l)}(\tilde{x}^{(l)}_i),
\end{align}

\vspace{-3 mm}
for $l = 1, \ldots, L$. To enable classification, a special classification token [CLS] is prepended to the input sequence. This token is designed to aggregate global sequence information throughout the transformer layers. After the final layer, the representation corresponding to the [CLS] token, denoted $x^{(L+1)}_{\texttt{[CLS]}}$, is passed through a linear classifier:

$$
\text{logits} = Wx^{(L+1)}_{\texttt{[CLS]}} \in \mathbb{R}^{C},
$$

where $W \in \mathbb{R}^{C \times d_{\text{model}}}$ is the classification weight matrix and $C$ is the number of target classes. A softmax function is then applied to the logits to produce a probability distribution $p = (p_1, p_2, \ldots, p_C)$ over the class labels.

\textbf{Vision Transformers (ViT).} \citep{dosovitskiy2020image} extend transformers to image classification by reshaping an image into a sequence of non-overlapping patches $p = (p_1, \ldots, p_k)$, where each $p_i \in \mathbb{R}^{P^2 \cdot C}$ corresponds to a flattened $P \times P$ patch with $C$ channels. These are projected into embeddings $x^{(1)}_i = \text{Linear}(p_i) + \text{PosEnc}(i)$, and then processed using the same transformer architecture as in NLP. A learnable class token is prepended to the patch sequence, and its output embedding after $L$ layers is used for classification.

\textbf{BERT.} \citep{devlin2019bert} is a bidirectional encoder-only transformer pretrained using masked language modeling (MLM) and next sentence prediction (NSP). Given a token sequence $t$, BERT computes contextualized embeddings using the same residual update pattern as Eq. (1). The [CLS] token, placed at the beginning of the input, is used as a summary representation for sentence-level classification tasks. The pretraining objective enables BERT to capture bidirectional dependencies, leading to strong performance across diverse NLP benchmarks.

\vspace{-3 mm}
\subsection{Extraction and Intervention of Steering Vectors}
\label{steering_vectors}

%%TODO: Modify this to fit our case, mention that we do class token in the selection and that we do abaltions for different methods of application.

\paragraph{Difference-in-means.}
To identify the ``bias direction'' in the model's residual stream activations, we compute the difference between the model's mean activations when run on groups with the same class but different confounding variables, which will be over or underrepresented in the class.
This technique, known as \emph{difference-in-means} \citep{belrose2023diffinmeans}, effectively isolates these key feature directions, as demonstrated in prior work \citep{panickssery2024steeringllama2contrastive, tigges2023linear}. For each layer $l \in [L]$ and post-instruction token position $i \in I$, we calculate the mean activation $\boldsymbol{\upmu}_i^{(l)}$ on images from $\mathcal{D}_{\text{overrepresented}}^{\text{(train)}}$ and $\boldsymbol{\upnu}_i^{(l)}$ for on images from $\mathcal{D}_{\text{underrepresented}}^{\text{(train)}}$:

\begin{table*}[ht!]
    \centering
        \caption{\textbf{Worst-group and Average-group Accuracy on Benchmark Datasets.} Comparison of our steering vector methods (full residual stream and best single layer ablation) with standard ERM and prior fairness approaches (FFR, GDRO) across Waterbirds, CelebA, UTKFace, and MultiNLI. ``Training Required?'' indicates whether the method requires retraining. $\dagger$ denotes results from prior work. Our methods are training-free and improve worst-group accuracy with minimal impact on average accuracy.}
    \label{tab:reduced-table}
    \small
    \begin{tabular}{c|c|c|cc}
    \toprule
    \multirow{2}{*}{Dataset} & \multirow{2}{*}{Method} & \multirow{2}{*}{\begin{tabular}[c]{@{}c@{}}Training \\ Required?\end{tabular}} & \multicolumn{2}{c}{Original Dataset} \\
    \cmidrule{4-5} 
     &  &  & Worst & Average \\
    \midrule
    \multirow{4}{*}{Waterbirds} 
        & ERM & - & $62.46$ & $89.43$ \\
        \cmidrule{2-5}
        & Full Residual Stream (Waterbirds class) & \XSolid &  $78.19$ & $93.18$ \\
        & Full Residual Stream (Landbirds class) & \XSolid &  $83.95$ & $92.49$ \\
        & Best Single Layer (Waterbirds Class) & \XSolid &  $75.9$ & $92.3$ \\
        & Best Single Layer (Landbirds Class) & \XSolid &  $76.5$ & $94.1$ \\
        \cmidrule{2-5}
        & FFR$\dagger$~\cite{qraitem2023fake} & \Checkmark &  $69.5$ & $84.0$ \\
        & GDRO$\dagger$~\cite{sagawa2019distributionally} & \Checkmark & $91.4$  & $93.5$ \\
    \midrule
    \multirow{4}{*}{CelebA} 
        & ERM & - &  $47.8$ & $94.9$ \\
        \cmidrule{2-5}
        & Full Residual Stream (Blond Hair)  & \XSolid &  $62.22$ & $93.47$ \\
        & Best Single Layer (Blond Hair)  & \XSolid &  $64.84$ & $94.15$ \\
        \cmidrule{2-5}
        & FFR$\dagger$~\cite{qraitem2023fake} & \Checkmark &  $68.9$ & $85.7$ \\
        & GDRO$\dagger$~\cite{sagawa2019distributionally} & \Checkmark  & $88.9$ & $92.9$ \\
    \midrule
    \multirow{4}{*}{UTKFace} 
        & ERM & - &  $74.3$ & $84.5$ \\
        \cmidrule{2-5}
        & Full Residual Stream(Male)  & \XSolid &  $50.98$ & $74.37$ \\
        & Full Residual Stream(Female)  & \XSolid &  $47.11$ & $79.16$ \\
        & Best Single Layer (Male)  & \XSolid &  $79.67$ & $88.20$ \\
        & Best Single Layer (Female)  & \XSolid &  $76.50$ & $86.09$ \\
        \cmidrule{2-5}
        & FFR$\dagger$~\cite{qraitem2023fake} & \Checkmark &  $67.4$ & $81.4$ \\
        & GDRO$\dagger$~\cite{sagawa2019distributionally} & \Checkmark & $81.6$ & $85.9$ \\
    \midrule
    \multirow{4}{*}{MultiNLI} 
        & ERM & - &  $47.8$ & $94.9$ \\
        \cmidrule{2-5}
        & Full Residual Stream (contradiction-negation)  & \XSolid & $77.67$ & $72.99$ \\
        & Best Single Layer (contradiction-negation)  & \XSolid & $69.9$ & $79.7$ \\
        \cmidrule{2-5}
        & FFR$\dagger$~\cite{qraitem2023fake} & \Checkmark & - & - \\
        & GDRO$\dagger$~\cite{sagawa2019distributionally} & \Checkmark  & $77.7$ & $81.4$ \\
    \bottomrule
    \end{tabular}
\end{table*}

% \begin{align}
% \boldsymbol{\upmu}_i^{(l)} = \frac{1}{|\mathcal{D}_{\text{overrepresented}}^{\text{(train)}}|} \sum_{\mathbf{t} \in \mathcal{D}_{\text{overrepresented}}^{\text{(train)}}} \mathbf{x}_i^{(l)}(\mathbf{t}) , \quad  
% \boldsymbol{\upnu}_i^{(l)} = \frac{1}{\lvert\mathcal{D}_{\text{underrepresented}}^{\text{(train)}}\rvert} \sum_{\mathbf{t} \in \mathcal{D}_{\text{underrepresented}}^{\text{(train)}}} \mathbf{x}_i^{(l)}(\mathbf{t}).
% \end{align}

\begin{align}
\boldsymbol{\upmu}_i^{(l)} &= \frac{1}{\left|\mathcal{D}_{\text{overrepresented}}^{\text{(train)}}\right|}
\sum_{\mathbf{t} \in \mathcal{D}_{\text{overrepresented}}^{\text{(train)}}}
\mathbf{x}_i^{(l)}(\mathbf{t}), \\
\boldsymbol{\upnu}_i^{(l)} &= \frac{1}{\left|\mathcal{D}_{\text{underrepresented}}^{\text{(train)}}\right|}
\sum_{\mathbf{t} \in \mathcal{D}_{\text{underrepresented}}^{\text{(train)}}}
\mathbf{x}_i^{(l)}(\mathbf{t}).
\end{align}

We then compute the difference-in-means vector $\mathbf{r}_i^{(l)} = \boldsymbol{\upmu}_i^{(l)} - \boldsymbol{\upnu}_i^{(l)}$.
%Note that each such vector is meaningful in both (1) its direction, which describes the direction that mean harmful and harmless activations differ along, and (2) its magnitude, which quantifies the distance between mean harmful and harmless activations.

\paragraph{Intervention with Steering Vector} We explore two primary methods to select the extracted activation means and then apply it to the residual stream during inference time. The two methods being full residual stream and single layer based interventions.

\paragraph{Directional ablation (Single Layer).}
To investigate the role of a direction $\hat{\mathbf{r}} \in \mathbb{R}^{d_{\text{model}}}$ in the model's computation, we can erase it from the model's representations using \emph{directional ablation} \citep{arditi2024refusal}.
Directional ablation ``zeroes out'' the component along $\hat{\mathbf{r}}$ for every residual stream activation $\mathbf{x} \in \mathbb{R}^{d_{\text{model}}}$:
\begin{align}
    \mathbf{x}' \leftarrow \mathbf{x} - \hat{\mathbf{r}} \hat{\mathbf{r}}^{\intercal} \mathbf{x}. \label{eq:projection}
\end{align}

Computing the difference-in-means vector $\mathbf{r}_{i}^{(l)}$ for position 0, the class token and layer $l \in [L]$ yields a set of L candidate vectors.
We specifically select the class token as it contains the global aggregate information used by the final classifier head, and thus most strongly impacts the final classification. We then select the single most effective vector $\mathbf{r}_{i^*}^{(l^*)}$ from this set by evaluating each candidate vector over a validation set $\mathcal{D}_{\text{val}}$.

We then perform the directional ablation operation at every activation $\mathbf{x}_{i}^{(l)}$ and $\tilde{\mathbf{x}}_{i}^{(l)}$, across all layers $l$ and all token positions $i$ using this selected vector ($\mathbb{R}^d_{\text{model}}$).
This effectively prevents the model from ever representing this direction in its residual stream.

\vspace{-4 mm}
\paragraph{Full Directional Ablation.}
We also explore \emph{full directional ablation}, which generalizes the concept of directional ablation \citep{arditi2024refusal}. Rather than removing a single direction from each residual stream vector, full directional ablation removes a specific direction at every layer and sequence position. Let $\mathbf{X} \in \mathbb{R}^{L \times T \times d_{\text{model}}}$ denote the residual stream activations, and let $\hat{\mathbf{R}} \in \mathbb{R}^{L \times T \times d_{\text{model}}}$ contain unit vectors specifying the direction to ablate at each layer and position. The ablated residual stream $\mathbf{X}'$ is given by:
\begin{align}
\mathbf{X}'[l, t] \leftarrow \mathbf{X}[l, t] - \hat{\mathbf{R}}[l, t] \left( \hat{\mathbf{R}}[l, t]^{\intercal} \mathbf{X}[l, t] \right),
\end{align}
for each layer $l \in \{1, \ldots, L\}$ and position $t \in \{1, \ldots, T\}$. This operation thus erases the component of each residual stream vector that lies in its corresponding direction, allowing for more fine grained access over the information being deleted.

\vspace{-3 mm}
\section{Results}
We evaluated the efficacy of model steering for debiasing in both Vision Transformers for 3 image datasets and BERT for 1 language dataset \citep{sagawa2019distributionally} along with exploration of different extraction and application techniques of the steering vector.

\subsection{Datasets and Experimental Setup}
\label{datasets}
% We evaluate on 3 image datasets, waterbirds with background bias, CelebA and UTKFace with facial attribute bias, and 1 language dataset MultiNLI with negative words correlation bias.
% Waterbirds consists of bird images with a binary label of waterbird 
% or landbird, with background bias, only a few waterbird images have land background and vice-versa \citep{sagawa2019distributionally}. CelebA  contains 202,599 face images; we use Blond Hair as the target attribute, which exhibits gender bias \citep{zhang2020celeba}. UTKFace \citep{zhang2017age} includes 20, 000 face images annotated with age, gender, and ethnicity; we use gender as the target and age as the bias attribute. Multi-NLI \citep{williams2017broad} contains 206,175 training examples with 1521 examples in the smallest group. It consists premise-hypothesis pairs, which are classified as one 3 labels $\mathcal{Y} = \{\text{entailed}, \text{neutral}, \text{contradictory}\}$ and
% spurious attributes $\mathcal{A} = \{\text{no negation}, \text{negation}\}$. This dataset contains spurious correlation between contradictions and the present of negation words \emph{nobody}, \emph{no}, \emph{never}, and \emph{nothing} \citep{gururangan2018annotation}.

We evaluate on four datasets exhibiting known biases:
\vspace{-3 mm}
\begin{itemize}
    \item \textbf{Waterbirds} \citep{sagawa2019distributionally}: Images of waterbirds or landbirds where labels are spuriously correlated with image backgrounds (e.g., waterbirds rarely appear on land backgrounds and vice-versa).
    \vspace{-1.5 mm}
    \item \textbf{CelebA} \citep{zhang2020celeba}: Face images (202,599 total) where the 'Blond Hair' attribute (target) exhibits significant gender bias. Hence, we evaluate only the bias vector for this class.
    \vspace{-1.5 mm}
    \item \textbf{UTKFace} \citep{zhang2017age}: Face images (20,000 total) using gender as the target attribute, with age as a known biasing attribute.
    \vspace{-1.5 mm}
    \item \textbf{MultiNLI} \citep{williams2017broad}: A textual entailment dataset (premise-hypothesis pairs with entailment, neutral, or contradiction labels) that exhibits a spurious correlation between contradiction labels and the presence of negation words (e.g., \emph{no}, \emph{never}, \emph{nothing}) \citep{gururangan2018annotation}.
\end{itemize}

\vspace{-3 mm}

Experiments utilize models pre-trained (e.g., on ImageNet) and subsequently fine-tuned on the respective dataset's training set. Steering vectors are then applied at inference time on the test set. We conduct extensive ablations on vector application methods, exploring single-layer, and simultaneous (all-at-once) interventions. Following established practice, we report worst-group (WGA) and average-group (AGA) accuracies.

\paragraph{Baselines}We compare against FFR~\citep{qraitem2023fake}, which uses synthetic data and a two-stage pipeline for fair representation learning, and GDRO~\citep{sagawa2019distributionally}, which optimizes worst-group accuracy during training (Table~\ref{tab:reduced-table}). Both require model retraining, unlike our post-hoc, training-free approach that achieves similar fairness results.

\subsection{Key Results}
Table~\ref{tab:reduced-table} reports WGA and AGA across datasets, evaluating the effectiveness of steering vectors and directional ablation in mitigating bias in both ViT and BERT models. We find that single-layer ablation applied to the [CLS] token yields the largest improvement in WGA with minimal impact on AGA because of minimal decrease in performance across the other groups, outperforming full ablation in terms of this tradeoff. Furthermore, the results achieved by our steering vector method are able to surpass performance of data-based methods like FFR\citep{qraitem2023fake} in case of UTKFace and Waterbirds. While reaching performance close to the state-of-the-art GDRO in the same. Overall, our results demonstrate that directional ablation of bias aligned vectors not only enhances fairness but also maintains strong performance across various datasets and models, making it a promising approach for improving fairness in transformer classifiers.

\vspace{-5 mm}
\paragraph{Minor Results} We observe that the steering vectors on layer-wise ablations are the most effective in middle-late layers which is conducive with other research in the field in VLMs and LLMs.

\vspace{-3 mm}
\section{Conclusion and Future Work}
\label{sec:conclusion}

This work introduced a training-free, inference-time method using directional ablation of "bias vectors"—derived from mean activation differences between skewed groups—to mitigate bias in transformer classifiers. Our experiments show this approach, especially single-layer [CLS] token interventions, significantly improves worst-group accuracy across diverse benchmarks with minimal impact on overall performance, offering a computationally inexpensive alternative to retraining. This research underscores the potential of direct activation manipulation as a computationally inexpensive and readily applicable tool for developing fairer and more reliable classification systems, extending the utility of steering vectors beyond generative AI into the classification domain.

Future work will focus on a rigorous mechanistic understanding of these interventions, enhancing steering vector techniques for scenarios like multiple interacting biases or dynamic application, evaluating performance in data-constrained or label-scarce settings, and broadening applicability to diverse model architectures and more complex bias scenarios, such as intersectionality. There is also scope of extending the idea of steering vectors for the classification domain to a more diverse set of problems like domain-shift etc.

\bibliography{example_paper}
\bibliographystyle{icml2025}

%%%%%%%%%%%%%%%%%%%%%%%%%%%%%%%%%%%%%%%%%%%%%%%%%%%%%%%%%%%%%%%%%%%%%%%%%%%%%%%
%%%%%%%%%%%%%%%%%%%%%%%%%%%%%%%%%%%%%%%%%%%%%%%%%%%%%%%%%%%%%%%%%%%%%%%%%%%%%%%
% APPENDIX
%%%%%%%%%%%%%%%%%%%%%%%%%%%%%%%%%%%%%%%%%%%%%%%%%%%%%%%%%%%%%%%%%%%%%%%%%%%%%%%
%%%%%%%%%%%%%%%%%%%%%%%%%%%%%%%%%%%%%%%%%%%%%%%%%%%%%%%%%%%%%%%%%%%%%%%%%%%%%%%
% \newpage
% \appendix
% \onecolumn
% \section{You \emph{can} have an appendix here.}

% You can have as much text here as you want. The main body must be at most $8$ pages long.
% For the final version, one more page can be added.
% If you want, you can use an appendix like this one.  

% The $\mathtt{\backslash onecolumn}$ command above can be kept in place if you prefer a one-column appendix, or can be removed if you prefer a two-column appendix.  Apart from this possible change, the style (font size, spacing, margins, page numbering, etc.) should be kept the same as the main body.
% %%%%%%%%%%%%%%%%%%%%%%%%%%%%%%%%%%%%%%%%%%%%%%%%%%%%%%%%%%%%%%%%%%%%%%%%%%%%%%%
% %%%%%%%%%%%%%%%%%%%%%%%%%%%%%%%%%%%%%%%%%%%%%%%%%%%%%%%%%%%%%%%%%%%%%%%%%%%%%%%

\end{document}